\documentclass{bmvc2k}

\title{When Text and Images Don't Mix: Bias-Correcting Language-Image Similarity Scores for Anomaly Detection}

\addauthor{Adam Goodge}{goodge_adam_david@i2r.a-star.edu.sg}{1}
\addauthor{Bryan Hooi}{dcsbhk@nus.edu.sg}{2}
\addauthor{Wee Siong Ng}{wsng@i2r.a-star.edu.sg}{1}

\addinstitution{
Institute for Infocomm Research \\ (I2R), A*STAR, Singapore
}
\addinstitution{
School of Computing, \\  National University of Singapore,  \\Singapore
}

\runninghead{GOODGE, ET AL.}{When Text and Images Don't Mix}

\usepackage{graphicx}
\usepackage{amsmath}
\usepackage{amsthm}
\usepackage{booktabs}
\usepackage{algorithm}
\usepackage{algorithmic}
\usepackage{subcaption}
\usepackage{amssymb}

\begin{document}

\maketitle

\begin{abstract}
Contrastive Language-Image Pre-training (CLIP) achieves remarkable performance in various downstream tasks through the alignment of image and text input embeddings and holds great promise for anomaly detection. However, our empirical experiments show that the embeddings of text inputs unexpectedly tightly cluster together, far away from image embeddings, contrary to the model's contrastive training objective to align image-text input pairs. We show that this phenomenon induces a `similarity bias' - in which false negative and false positive errors occur due to bias in the similarities between images and the normal label text embeddings. To address this bias, we propose a novel methodology called BLISS which directly accounts for this similarity bias through the use of an auxiliary, external set of text inputs. BLISS is simple, it does not require strong inductive biases about anomalous behaviour nor an expensive training process, and it significantly outperforms baseline methods on benchmark image datasets, even when access to normal data is extremely limited.
\end{abstract}

\section{Introduction}

Anomaly detection (AD) is an important task in many vision-related applications, such as medical diagnosis and industrial defect detection. Neural networks are trained to embed input images into a latent space where anomalies are more easily detected. Vision-language models (VLM), which embed both image and textual inputs, have surged in popularity recently due to their flexibility and strong performance in various downstream tasks. CLIP \cite{radford2021learning} is particularly noteworthy; it is contrastively trained to maximise the cosine similarity between the latent embeddings of image-text caption pairs and minimise it between non-pairs. A query image should have high similarity to text inputs of its class label, and low similarity to unrelated text inputs, which is exploited for predicting class membership.

In this work, we conduct empirical experiments to examine the latent space learnt by CLIP. We find that, contrary to its contrastive training objective, all text inputs are highly clustered together despite their large differences in semantic content. The result is that textual class labels exhibit significantly higher similarity to other unrelated text inputs than they do to the relevant images, which we call the `text clustering effect'. This phenomenon violates expected behaviour, and we find that it also impacts anomaly detection performance by inducing bias in anomaly scores, which we call the `similarity bias'. We show in our experiments that this bias is exposed in the similarities of different images to general text embeddings, beyond those of the restricted normal classes. 

We address this issue by proposing a novel methodology called `\textbf{B}ias-corrected \textbf{L}anguage \textbf{I}mage \textbf{S}imilarity \textbf{S}coring' (BLISS), which addresses similarity bias by directly accounting for the similarity of test images to general text embeddings. BLISS is simple, it avoids strong inductive biases about anomalous behaviour, and it is highly efficient; it is entirely inference-based and requires no model training. We show in comprehensive experiments that BLISS outperforms comparable baselines on key benchmark datasets and it exhibits strong robustness to different modifications and problem settings. Our code will be made available online upon publication.

In summary, the main contributions of this paper are:
\begin{enumerate}
    \item We identify a `text clustering effect' in the latent space of CLIP and analyse how this effect induces a `similarity bias' in anomaly scoring.
    \vspace{-0.2cm}
    \item We propose a novel methodology called BLISS which accounts for this bias in its anomaly scoring and outperforms competing methods.
    \vspace{-0.2cm}
    \item We examine the performance of different components of our methodology and demonstrate its effectiveness in several experimental settings.
\end{enumerate}

\section{Related Work}

There are two types of tasks commonly referred to as anomaly detection. In the first, normal data is characterized by its belonging to a normal class while anomalies are any samples that do not belong to a normal class. In this setting, existing methods exploit properties such as the distance to normal samples \cite{angiulli2002fast,goodge2022lunar} or a normal hyper-sphere \cite{ruff2019deep} in the latent space to detect anomalies. Some methods train models to perform auxiliary tasks, such as reconstruction \cite{an2015variational,chen2018autoencoder,zhou2017anomaly}, generation \cite{zenati2018efficient,akcay2018ganomaly,schlegl2017unsupervised,schlegl2019f} or classification \cite{golan2018deep,hendrycks2019using,tack2020csi} instead, expecting the model to generalise to other normal samples in the test set but not to anomalies. In this work, we refer to this task as 'semantic anomaly detection', and it is highly related to the similar `out-of-distribution detection' task. On the other hand, the second type of anomaly detection task characterizes anomalies as small aberrations from otherwise highly uniform normal data. This setting is most often seen in industrial defect detection \cite{li2021cutpaste,perera2019ocgan,roth2022towards,defard2021padim}, for which several methods exploiting CLIP's multi-modal capabilities have emerged in recent years \cite{jeong2023winclip,li2024promptad}. However, these methods typically require some level of prior knowledge about anomalous behaviour, which limits their applicability in settings where anomalies can result from unknown and unpredictable sources. Indeed, even the strongest methods in industrial defect detection tend to perform poorly in semantic AD \cite{liu2023simplenet}.

In this work, we focus on the semantic AD task, making no assumptions about the nature of anomalies in the test set. Existing zero-shot CLIP-based methods \cite{esmaeilpour2022zero,liznerski2022exposing,ming2022delving} do not exploit information from labelled normal images. However, in practical settings, it is widely acknowledged that prior examples of normal data are often accessible. These examples are highly valuable in capturing the learned notion of normality by the model, and our methodology is designed to exploit this information not to fine-tune the model parameters but to enhance the anomaly scoring process at test-time.

\section{Motivation}\label{sec:motivation}

In this section, we examine the latent space of CLIP and find what we call the `text clustering effect'. We show the impact of this phenomenon on anomaly scoring, namely that it induces a `similarity bias' which causes prediction errors. We begin by defining the semantic anomaly detection task of interest in this work.

\subsection{Problem Statement}

We have labelled normal samples $\mathbf{X}^{(train)} = \{(\mathbf{x}_{1}^{(train)}, y_1^{(train)}), ..., (\mathbf{x}_{n}^{(train)},y_n^{(train)})\}$, where  $y_i^{(train)}$ is one of the normal classes with text labels $\mathcal{C} = \{\mathcal{C}_1, ..., \mathcal{C}_N \}$. We also have a set of unlabelled test images $\mathbf{X} = \{\mathbf{x}_{1}, ..., \mathbf{x}_{m}\}$, each of which may be normal (belonging to a class in $\mathcal{C}$) or anomalous. Our goal is to define a function $s:\mathbf{X} \to \mathbb{R}$ which maps an input image $\mathbf{x}$ to an anomaly score $s(\mathbf{x})$ which is low for normal images and high for anomalies.
  
\subsection{Text Clustering Effect}

\begin{figure}
\centering
\begin{tabular}{cc}
\includegraphics[width=5cm,height = 5cm]{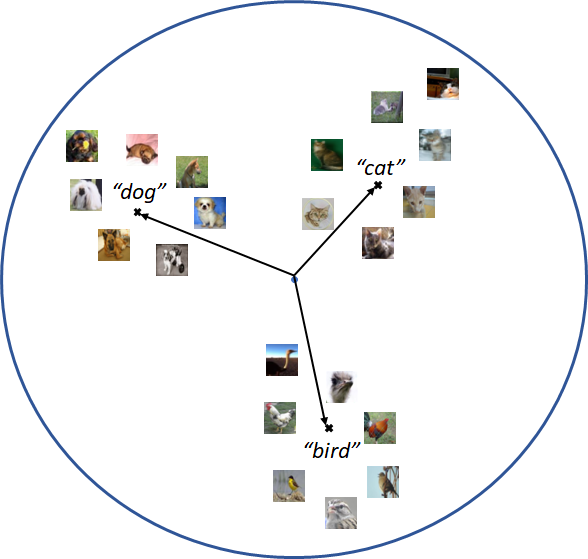}&
\includegraphics[width=5cm, height = 5cm]{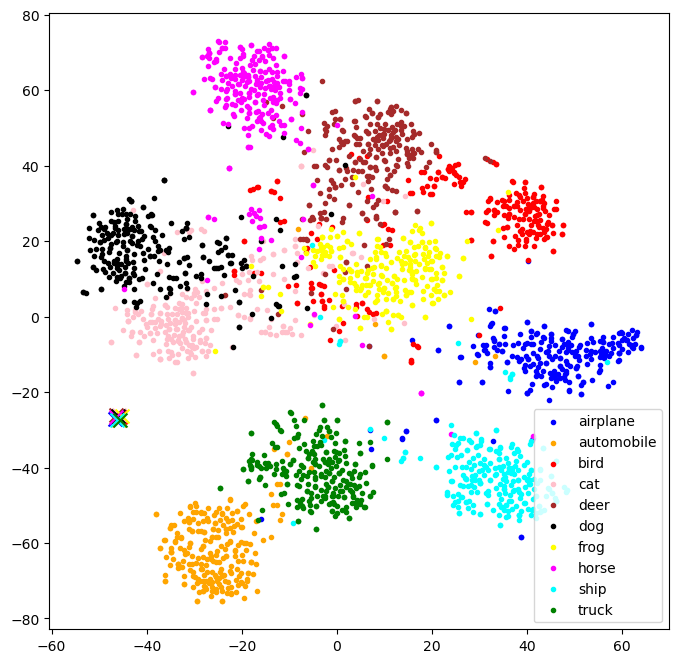}\\
(a)&(b) \\
\end{tabular}
\vspace{0.2cm}
\caption{(a) The na\"ive expectation of CLIP's latent space based on its contrastive objective. Text labels which are not semantically similar are separated from each other and the corresponding images cluster around them. (b) t-SNE projections of the true embeddings learnt by CLIP. Text labels (crosses) are highly clustered together, far away from images (dots).}
\label{fig:teaser}
\end{figure}

Based on its contrastive training objective, a reasonable expectation of the latent space learnt by CLIP is illustrated in Figure \ref{fig:teaser}(a). Using CIFAR-10 \cite{krizhevsky2009learning} for demonstration, the class labels \{`\textit{dog}', `\textit{cat}' and `\textit{bird}'\} are distanced from each other (exaggerated for clarity) as they are semantically distinct concepts, while their associated images tightly cluster around their paired labels. In Figure \ref{fig:teaser}(b), we show that this expectation is not met in reality. We plot the t-SNE projections \cite{JMLR:v9:vandermaaten08a} of the real embeddings of CIFAR-10 class labels (crosses) and images (dots). Following previous work, class labels are converted into text prompts with the template: ``\textit{This is a photo of a} $\{$\textit{class label}$\}$". All embeddings are normalized to the unit hyper-sphere.

We immediately see that the text and image embeddings occupy highly separated regions of the latent space. The images embeddings are broadly clustered by class, and these clusters are dispersed throughout the latent space. On the other hand, the text embeddings are tightly clustered together (they all visually overlap at approximately (-45,-25) in the plot). This is despite the fact that most labels are not at all related, for example the ``\textit{airplane}" and ``\textit{cat}" class labels. We call this phenomenon the `text clustering effect'.

\begin{figure}[t]
    \centering
    \includegraphics[width = 8cm, height = 3cm]{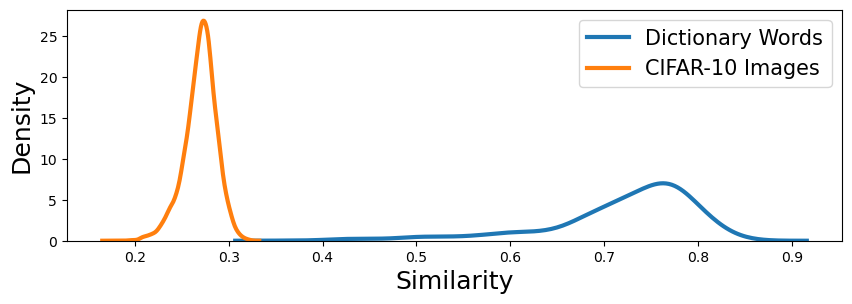}
    \vspace{0.2cm}
    \caption{Average cosine similarities of the normalized CLIP embeddings of CIFAR-10 text class labels to the dictionary (blue), and to their associated CIFAR-10 image embeddings (orange).}
    \label{avg_sim_kde}
\end{figure}

Figure \ref{avg_sim_kde} supports this observation. The orange line shows the distribution of cosine similarities between each CIFAR-10 image and its correct class label, which we see peaks around 0.25. On the other hand, the blue line shows the average similarity between the same class labels and a set of general purpose text inputs (we use ImageNet \cite{deng2009imagenet} class labels), which we refer to as the `dictionary'. As the dictionary entries covers a very large and broad range of concepts, the vast majority of entries should be unrelated to any given class label, and we may therefore expect the average similarity of each class label to the dictionary to be relatively low. However, we see that it is actually vastly higher than it is for the image inputs, peaking at around 0.75. This means that the class label text inputs are significantly more similar to other text inputs regardless of their content than they are to the images they are supposed to describe. This is despite the fact that CLIP was directly trained to maximise the similarity between images and their related text captions. Furthermore similar plots in the supplementary material with embeddings from another VLM, BLIP \cite{li2022blip}, shows that this text clustering effect is not unique to CLIP but a common feature of vision-language modelling.

\subsection{Similarity Bias in Anomaly Scoring}\label{sec:sim_bias}

\begin{figure}
\centering
\includegraphics[width = 8cm, height = 3cm]{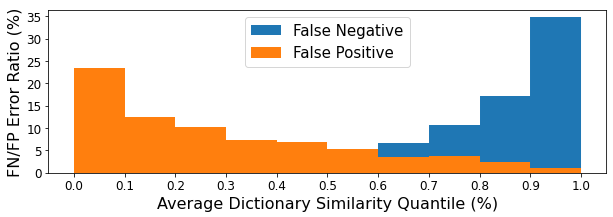}
\vspace{0.2cm}
\caption{Proportion of false negative (blue) and false positive (orange) errors in each quantile of samples sorted by average similarity to the dictionary.}
\label{fig:fp_fn_quantiles}
\end{figure}
 
A simple approach to anomaly scoring would be to measure the similarity of a query image to the normal class label(s). A higher similarity suggests the sample is normal, while a lower similarity indicates an anomaly. However, the text clustering effect identified above means that normal class labels are tightly clustered with unrelated text labels, potentially including labels of anomalies. We now investigate the impact of this finding on AD performance.

To this end, we measure the average similarity of CIFAR-10 images to the dictionary mentioned earlier. As the dictionary is large and conceptually diverse, this average similarity can be interpreted as a measure of the similarity of an image to general text embeddings, rather than to  their specific class label. In theory, any given image should be meaningfully related to only a few entries and unrelated to the vast majority of them, meaning the average dictionary similarity should be not be a very informative measure of an image's anomaly status. However, in Figure \ref{fig:fp_fn_quantiles}, we show that the distribution of errors from this anomaly scoring approach is highly correlated with average dictionary similarity. In particular, false negative errors (anomalies classified as normal) are much more frequent amongst images with high average dictionary similarity (right side of the plot). Conversely, false positive errors (normal samples classified as anomalies) are much more frequent in images with low average dictionary similarity (left side of the plot). In other words, anomalies that are relatively more similar to general text embeddings are also more similar to the normal class labels, which makes them appear normal, and vice versa for normal samples that are relatively less similar. We call this phenomenon the `similarity bias', as the discrepancy in similarities of different images to general text embeddings causes bias in their anomaly scores and degrades AD performance. We aim to address this similarity bias in our proposed methodology by directly accounting for similarity to general text embeddings.

\section{Methodology}

BLISS consists of two components; an \textbf{internal class score} and \textbf{external text score}. We rely on the fixed, pre-trained CLIP backbone model with image encoder $\mathcal{I}$ and text encoder $\mathcal{T}$. We firstly obtain all of the embeddings of labelled normal images using the image encoder and store them in a memory bank:
\begin{align}
    & \mathbf{Z}^{(train)} = \{\mathbf{z}^{(train)}_{1}, ..., \mathbf{z}^{(train)}_{n}\}, \ \text{where} \ \mathbf{z}^{(train)}_{i} = \mathcal{I}(\mathbf{x}^{(train)}_i),
\end{align}
Note that the encoders are fixed and we do not fine-tune any model parameters. For a given unlabelled test image, $\mathbf{x}$, we similarly find its own embedding:
\begin{align}
    \mathbf{z} = \mathcal{I}(\mathbf{x}).
\end{align}
We now define the two components, starting with the internal class score.

\subsection{Internal Class Score}\label{internal class_score}
 
The internal class score measures the normality of the test image based on its similarity to the normal class label(s). We obtain all of the text embeddings from the normal class labels with the fixed CLIP text encoder:
\begin{align}
    & \mathbf{C} = \{\mathbf{C}_{1},..,\mathbf{C}_{N}\}, \ \text{where} \ \mathbf{C}_{i} = \mathcal{T}(\mathcal{C}_i).
\end{align}
For each normal class, we collect all of the normal samples belonging to that class and compute the mean ($\mathsf{mean}$) and standard deviation ($\mathsf{std}$) of their similarities ($sim$) to $\mathbf{C}_i$:
\begin{align}
    & \mu_i = \underset{y^{(train)}_i = \ \mathcal{C}_i}{\mathsf{mean}}sim(\mathbf{z}^{(train)}_i, \mathbf{C}_i), \ \ \ \ \ \ 
    \ \  \ \ \sigma_i = \underset{y^{(train)}_i = \ \mathcal{C}_i}{\mathsf{std}} sim(\mathbf{z}^{(train)}_i, \mathbf{C}_i)
\end{align}
The internal class score of $\mathbf{x}$ is then the normalized similarity $sim(\mathbf{z}, \mathbf{C}_i)$ with respect to these statistics:
\begin{align}\label{eq:calibration}
    IC(\mathbf{x},\mathcal{C}_i) = -\frac{sim(\mathbf{z},\mathbf{C}_i)-\mu_{i}}{\sigma_{i}+\epsilon},
\end{align}
where $\epsilon$ is a small constant to avoid division by zero. If $\mathbf{x}$ belongs to $\mathcal{C}_i$, then it is a normal sample and $sim(\mathbf{z},\mathcal{S}_i)$ should be high and therefore $IC(\mathbf{x}_{test},\mathcal{C}_i)$ should be low. Conversely, the opposite should be true if $\mathbf{x}$ does not belong to $\mathcal{C}_i$.

\subsection{External Text Score}\label{external text_score}

The external text score is designed to address the similarity bias. It measures the similarity of images to the dictionary representing general text inputs. We embed each entry in the dictionary $\mathcal{D} = \{\mathcal{D}_1, ..., \mathcal{D}_t\}$ with the CLIP text encoder:
\begin{align}
     & \mathbf{D} := \{\mathbf{D}_{1},..,\mathbf{D}_{t}\}, \ \text{where} \ \mathbf{D}_{i} = \mathcal{T}(\mathcal{D}_i).
\end{align}
For the given test image $\mathbf{x}$, with embedding $\mathbf{z}$, we denote its top $K$ closest matches with the highest similarity in $\mathbf{D}$ as $\mathbf{D}^*$:
\begin{equation}\label{eq:topk}
    \mathbf{D}^* = \mathsf{topK}\{sim(\mathbf{z},\mathbf{D}_i):\mathbf{D}_i \in \mathbf{D}\}
\end{equation}
Note that the elements of $\mathbf{D}^*$ are specific to the given test sample. As before, we find the mean and standard deviation statistics from the labelled images from each normal class to each $\mathbf{d}^* \in \mathbf{D}^*$:
\begin{align}
    & \mu_{i}(\mathbf{d}^*) = \underset{y^{(train)}_i = \mathcal{C}_i}{\mathsf{mean}}sim(\mathbf{z}^{(train)}_i,\mathbf{d}^*)  \ \ \ \ \ \ 
    \ \  \ \  \sigma_{i}(\mathbf{d}^*) =  \underset{y^{(train)}_i = \mathcal{C}_i}{\mathsf{std}}sim(\mathbf{z}^{(train)}_i,\mathbf{d}^*).
\end{align}
The external text score is then similarity computed as the mean of $sim(\mathbf{z},\mathbf{d}^*)$ over all $\mathbf{d}^* \in \mathbf{D}^*$, normalized by these statistics:
\begin{align}
    ET(\mathbf{x},\mathcal{C}_i) = \underset{\mathbf{d}^* \in \ \mathbf{D}^*}{\mathsf{mean}}\frac{sim(\mathbf{z},\mathbf{d}^*)-\mu_i(\mathbf{d}^*)}{\sigma_{_i}(\mathbf{d}^*) + \epsilon},
\end{align}

Intuitively, $\mu_i(\mathbf{d}^*)$ and $\sigma_{_i}(\mathbf{d}^*)$ capture the learned notion of normality using statistics from each normal class. After correcting for these statistics, test samples with high external text scores are those with high similarity to the dictionary words. From our observation of similarity bias, such samples tend to appear normal in terms of internal class score, and thus have a higher chance of being false negatives. The external text score corrects this by `subtracting away' this bias, leaving behind a more meaningful view of the similarity between an image and the normal class labels. As our method only requires forward passes through the fixed backbone model, it is light-weight and efficient to compute even with a very large dictionary size. 

\subsection{BLISS Score}
We take a linear combination of the internal class and external text scores, regulated by the hyper-parameter $\lambda$:
\begin{align}\label{eq:single_score}
s(\mathbf{x},\mathcal{C}_i) = IC(\mathbf{x}, \mathcal{C}_i) + \lambda ET(\mathbf{x},\mathcal{C}_i).
\end{align}
In the case of multiple normal classes, i.e. $|\mathcal{C}|>1$, we repeat scoring over every $\mathcal{C}_i \in \mathcal{C}$ and take the minimum as the final anomaly score:
\begin{align}\label{BLISS_score}
s(\mathbf{x}) = \min_{\mathcal{C}_i \in \mathcal{C}} s(\mathbf{x},\mathcal{C}_i).
\end{align}
In summary, we \emph{jointly} consider similarity not only to the normal class labels (internal) but also to general text embeddings (external), both standardized to account for the learned distribution of normal data.

\section{Experiments}

We now perform experiments to answer the following questions about our methodology:

\noindent \textbf{RQ1 (Performance):} Does BLISS outperform baseline methods in anomaly detection on benchmark datasets? \\
\textbf{RQ2: (Ablation Study)} How do the components of BLISS perform in different experimental settings?

\paragraph{Datasets}\label{sec:Datasets}

We use the most popular datasets in semantic AD: CIFAR-10, CIFAR-100 \cite{krizhevsky2009learning} and TinyImageNet \cite{le2015tiny}. For CIFAR-10, we partition the 10 classes into 1/6/9 normal classes and 9/4/1 anomaly classes in different experiments. For 1 normal class and 9 normal classes, we run 10 trials - one for each permutation of normal and anomaly class splits. For 6 normal classes, and for CIFAR-100 and TinyImageNet experiments, we run 5 trials with the same class splits as \cite{esmaeilpour2022zero}.

\paragraph{Model Setup}

We use the publicly available pre-trained ViT-B/16 \cite{dosovitskiy2020image} CLIP model and do not fine-tune any model parameters. Images are resized, center-cropped and normalized. We use ImageNet class labels as the external dictionary. Classes containing multiple labels, e.g. ``\textit{goldfish, Carassius auratus}" are treated as separate entries ``\textit{goldfish}" and ``\textit{Carassius auratus}", resulting in 1850 dictionary entries. The text prompt ``\textit{This is a photo of a} $\{$\textit{class label}$\}$" is formulated for each class label and dictionary entry. We try different prompt formulations in the supplementary material and find little effect on performance. For both modalities, we normalized their 512-dimensional embeddings to the unit hyper-sphere. We set $K=10$ for the external text score and $\lambda=0.5$ to approximately match the range of the two scores and we do not conduct hyper-parameter search for optimal performance.

\paragraph{Baselines}
We copy the results from \cite{esmaeilpour2022zero} of their method ZOC, CSI \cite{tack2020csi} and CAC \cite{millerclass}, for which some experiments are missing (marked by -). We also implement our own baseline methods as follows. To measure the role of multi-modal learning, we use features from the image-only ViT-B/16 model pre-trained on ImageNet-21k from HuggingFace \cite{wu2020visual} with AD methods: DN2 ($K$ = 5) \cite{bergman2020deep}, LUNAR \cite{goodge2022lunar} and Gaussian mixture model (GMM). DINO-FT uses KNN with the same ViT architecture but trained via DINO self-distillation \cite{reiss2022anomaly}. Out of CLIP-based methods, CAC, DN2, LUNAR and GMM are the same methods as above but using CLIP embeddings. MCM \cite{ming2022delving} and BCE-CL \cite{liznerski2022exposing} are two contemporary methods which also use text embeddings for anomaly scores, though note that these are zero-shot and do not utilise labelled normal samples. There is no result for MCM in CIFAR-10 with one normal class as the method requires more than one normal class. We present their performance primarily for reference. Biased-CLIP is the internal class score alone, which is vulnerable to similarity bias. For fairness, we choose not to consider methods which (i) require prior knowledge about anomalies or (ii) fine-tune the parameters of a pre-trained model. We also do not consider older baselines due to poor performance (<90 AUROC on CIFAR-10 one-class experiments). We use in PyTorch in Windows with an Nvidia GeForce RTX 2080 Ti GPU.

\subsection{RQ1: (Performance)} 

\begin{table*}[t!]
\setlength{\tabcolsep}{4.5pt}
\centering
\begin{tabular}{llllll}
\toprule
Dataset & CIFAR-10 & CIFAR-10 & CIFAR-10 & CIFAR-100 & TinyImageNet \\
$|\mathcal{C}|$ & 1 class & 6 classes & 9 classes & 20 classes & 20 classes\\
\midrule
CAC \cite{millerclass} & - & $80.1 \pm 3.0$ & $75.4 \pm 6.0$ & $76.1 \pm 0.7$ & $76.0 \pm 1.5$ \\
CSI \cite{tack2020csi} & - & $87.0 \pm 4.0$ & - & $80.4 \pm 1.0$ & $76.9 \pm 1.2$ \\
DN2 \cite{bergman2020deep}& $97.2 \pm 1.2$ & $89.3 \pm 3.9$ & $86.0 \pm 8.8$ & $83.5 \pm 0.8$ & $84.7 \pm 1.6$ \\
LUNAR\cite{goodge2022lunar} & $96.7 \pm 1.6$ & $88.5 \pm 4.2$ & $85.8 \pm 8.9$ & $82.9 \pm 1.0$ & $83.4 \pm 1.7$ \\
GMM & $97.7 \pm 0.9$ & $91.7 \pm 3.8$ & $88.0 \pm 11.2$ & $86.2 \pm 0.6$ & $89.0 \pm 1.4$ \\
DINO-FT \cite{reiss2022anomaly} & $98.7 \pm 0.9$ & $94.0 \pm 3.0$ & $89.5 \pm 8.3$ & $88.1 \pm 0.7$ & $90.2 \pm 1.3$ \\
\midrule
\multicolumn{6}{l}{\textbf{CLIP-based Methods}} \\
\midrule
CAC \cite{millerclass} & - & 89.3 $\pm$ 2.0 & - & 83.5 $\pm$ 1.2 & 84.6 $\pm$ 1.7 \\
DN2 \cite{bergman2020deep}& $96.2 \pm 1.9$ & $87.6 \pm 5.2$ & $83.7 \pm 11.5$ & $78.2 \pm 2.8$ & $79.4 \pm 2.4$ \\
LUNAR \cite{goodge2022lunar}& $96.3 \pm 2.1$ & $87.7 \pm 3.9$ & $85.0 \pm 10.6$ & $79.0 \pm 2.9$ & $80.4 \pm 2.2$ \\
GMM & $97.8 \pm 0.9$ & $91.7 \pm 2.6$ & $89.5 \pm 4.9$ & $77.3 \pm 2.8$ & $79.8 \pm 2.4$ \\
ZOC \cite{esmaeilpour2022zero}& - & $93.0 \pm 1.7$ & - & $82.1 \pm 2.1$ & $84.6 \pm 1.0$ \\
MCM \cite{ming2022delving} & - & 89.8 $\pm$ 6.8 & 87.4 $\pm$ 13.6 & 82.7 $\pm 1.2 $ & 82.6 $\pm$ 1.2 \\
BCE-CL \cite{liznerski2022exposing} &  98.6 $\pm$ 0.8 & 93.8 $\pm$ 3.5 & 91.7 $\pm$ 6.1 & 85.6 $\pm$ 0.5 & 87.4 $\pm$ 1.4\\
Biased-CLIP & $97.8 \pm 0.9$ & $93.2 \pm 2.0$ & $91.2 \pm 3.3$ & $82.6 \pm 2.6$ & $81.3 \pm 6.2$ \\
BLISS & $\mathbf{99.1 \pm 0.4}$ & $\mathbf{97.3 \pm 0.8}^*$ & $\mathbf{96.0 \pm 1.9}^*$ & $\mathbf{89.4 \pm 0.6}^*$ & $\mathbf{91.1 \pm 1.8}$ \\
\bottomrule
\end{tabular}
\vspace{0.3cm}
\caption{AUROC performance of BLISS compared against baselines on several datasets. We show the mean and standard deviation  of scores across trials detailed in section \ref{sec:Datasets}. The highest score for each dataset is highlighted in bold. $^*$ indicates statistical significance at $p=0.05$ against the next best performing baseline with the one-sided T-test.}
\label{tab:main_results}
\end{table*}

Table \ref{tab:main_results} shows the mean and standard deviation in AUROC scores. We use the AUROC metric to avoid choosing an anomaly score threshold. We see that BLISS outperforms all baselines in all datasets and often by a significant margin, a result of accounting for the similarity bias as well as exploiting normality statistics from labelled normal samples.

As TinyImageNet is a subset of ImageNet, all of its class labels are also present in the dictionary. This could be seen as a violation of the assumption of no prior knowledge, therefore we also measured performance after removing all TinyImageNet labels from the dictionary. We find only a small drop in performance from 91.1 to 90.4 AUROC, and this still outperforms all baselines. This issue does not affect CIFAR-10 or CIFAR-100 experiments.

\begin{figure}
    \centering
    \includegraphics[width=0.5\linewidth]{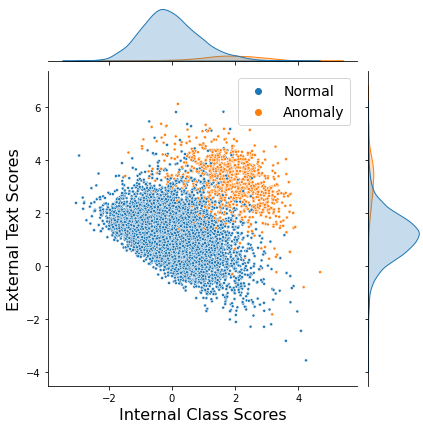}
    \vspace{0.1cm}
    \caption{Scatter plot of internal class scores (x-axis) vs. external text scores (y-axis) for normal (blue) and anomaly (orange) samples. }
    \label{fig:scatter}
\end{figure}

Figure \ref{fig:scatter} shows the internal class and external text scores of both normal (blue) and anomalous (orange) samples independently. Using only the internal class information for scoring is akin to discriminating anomalies based only on the horizontal axis. We see that the
optimal decision boundary is closer to a diagonal line, which is achieved through combining the information from both internal class and external text scores.

\subsection{RQ2: (Ablation Study)}

\paragraph{Weight hyper-parameter} Table \ref{tab:lambda} shows the performance of BLISS as $\lambda$ in (\ref{eq:single_score}) is varied. A higher value gives more weighting to the external text score. As the internal class score is generally smaller in magnitude than the external text score, we heuristically set $\lambda = 0.5$ to approximately equalise their weight. However, we see that $\lambda = 0.75$ is better for performance across all datasets. Overall, performance is robust within a reasonable range of $\lambda$.

\begin{table}[h!]
    \centering
    \setlength{\tabcolsep}{6pt}
    \begin{tabular}{lccccc}
    \toprule
    & CIFAR-10 & CIFAR-10 & CIFAR-10 & CIFAR-100 &  TinyImageNet  \\ 
     $\lambda$ & 1 class & 6 classes & 9 classes & 20 classes & 20 classes\\
    \midrule
      0.1 & 98.34  & 94.66  & 92.90  & 84.94 & 	84.85 \\ 
      0.25 & 98.82  & 96.12 & 94.69  &87.4 &	88.42 \\ 
      0.5 & 99.14  & 97.27  & 96.01 &
      89.43  &	91.13  \\ 
      0.75 & \textbf{99.18} & \textbf{97.62} & \textbf{96.45} & \textbf{89.73 } &	\textbf{91.92 }\\ 
      1 & 99.10  & 97.59 & 96.38  & 89.09  & 91.88  \\ 
      2 & 98.32  & 96.20  & 94.34  & 84.13 & 	89.52 \\ 
    \bottomrule
    \end{tabular}
        \vspace{0.2cm}
    \caption{Mean AUC scores of BLISS for different values of $\lambda$ in weighting the two scores. We see that the optimal performance is achieved around $\lambda = 0.75$ across all datasets, but performance is highly robust for a large range.}
    \label{tab:lambda}
\end{table}

\paragraph{Performance with other backbone VLMs}
\cite{liang2022mind} shows the embedding gap between different modalities in several multi-modal models. We now examine whether BLISS improves AD performance with other backbone VLMs. In Table \ref{tab:blip_scores}, we show the performance of the internal class score without normalization using labelled normal samples (``w/o (\ref{eq:calibration})"), with normalization (``w (\ref{eq:calibration})"), and our full BLISS score. We see that BLISS achieves similar performance gains over the internal class score alone with BLIP \cite{li2022blip} and SLIP \cite{mu2022slip} backbone VLMs too, proving the generalisation of our methodology.

\begin{table}
    \centering
        \setlength{\tabcolsep}{4pt}
    \begin{tabular}{lccc|ccc|ccc}
    \toprule
     & \multicolumn{3}{c}{CLIP} & \multicolumn{3}{c}{BLIP} & \multicolumn{3}{c}{SLIP}  \\ 
    & w/o (\ref{eq:calibration}) & w (\ref{eq:calibration}) & BLISS & w/o (\ref{eq:calibration}) & w (\ref{eq:calibration}) & BLISS & w/o (\ref{eq:calibration}) & w (\ref{eq:calibration}) & BLISS \\ 
    \midrule
    1 class & 94.69 & 97.80 & \textbf{99.14} & 96.47 & 96.47 & \textbf{97.93} & 96.08 & 96.08 & \textbf{97.96} \\ 
    6 classes & 86.17 & 93.19 & \textbf{97.27} & 86.50 & 89.86 & \textbf{94.20} & 84.96 & 88.46 & \textbf{92.33} \\ 
    9 classes & 82.16 & 91.20 & \textbf{96.02} & 85.41 & 88.13 & \textbf{92.60} & 84.25 & 86.28 & \textbf{90.56} \\ 
    \bottomrule
    \end{tabular}
    \vspace{0.2cm}
    \caption{Mean AUROC performance of the internal class score without calibration (Eq. \ref{eq:calibration}), (ii) the internal class score with calibration, and (iii) the full BLISS score including external text score, using different backbone VLMs.}
    \label{tab:blip_scores}
\end{table}

Our method diverges from recent zero-shot CLIP-based methods in its use of labelled normal samples. This could be seen as a limitation in settings where labelled normal samples are difficult to obtain. In the supplementary material, we measure the impact of restricting the number of labelled samples. We find that BLISS performance remains highly robust and out-performs the other methods even with as few as one sample per normal class, proving its suitability for few-shot anomaly detection. This shows the importance of correcting for similarity bias. We also find it is robust to changing $K$ in (\ref{eq:topk}). We also test BLISS with other dictionary sources and find that performance is aided by larger and broader dictionaries. Finally, we show that BLISS is highly effective in challenging cases by specially choosing semantically close normal vs. anomaly classes (e.g. horse vs. deer classes). We also measure the false positive rate at 95\% recall (FPR95) performance, which is an important metric for AD in practice, and find BLISS also improves performance in this metric.

\section{Conclusion}

In this paper, we identified a peculiarity of the CLIP latent space that all text embeddings are highly clustered together, away from their associated images. This hinders anomaly detection performance, and may also have important implications for other downstream tasks. We propose a novel anomaly scoring approach called BLISS which addresses this bias by measuring similarities against a large, external source of text inputs. BLISS is fast, anomaly-agnostic and achieves state-of-the-art performance on benchmark datasets. More broadly, our findings highlight the need for greater understanding of multi-modal learning and the need to address their unexpected characteristics; a goal of fundamental importance.

\bibliography{ref}
\end{document}